# Improving genetic risk prediction across diverse population by disentangling ancestry representations

Prashnna K Gyawali[1], Yann Le Guen[1,2], Xiaoxia Liu[1], Hua Tang[3], James Zou[4*], Zihuai He[1,5*]

[1]Department of Neurology and Neurological Sciences, Stanford University, Stanford, CA, USA
[2] Institut du Cerveau - Paris Brain Institute - ICM, Paris, France
[3]Department of Genetics, Stanford University, Stanford, CA, USA
[4]Department of Biomedical Data Science, Stanford University, Stanford, CA, USA
[5]Quantitative Sciences Unit, Department of Medicine (Biomedical Informatics Research), Stanford University, Stanford, CA, USA

* co-corresponding authors

**One Sentence Summary:** Deep learning model for disentangling ancestry representations and related population bias improves genetic risk prediction across diverse population.

**Abstract:** Risk prediction models using genetic data have seen increasing traction in genomics. However, most of the polygenic risk models were developed using data from participants with similar (mostly European) ancestry. This can lead to biases in the risk predictors resulting in poor generalization when applied to minority populations and admixed individuals such as African Americans. To address this bias, largely due to the prediction models being confounded by the underlying population structure, we propose a novel deep-learning framework that leverages data from diverse population and disentangles ancestry from the phenotype-relevant information in its representation. The ancestry disentangled representation can be used to build risk predictors that perform better across minority populations. We applied the proposed method to the analysis of Alzheimer's disease genetics. Comparing with standard linear and nonlinear risk prediction methods, the proposed method substantially improves risk prediction in minority populations, particularly for admixed individuals.

## INTRODUCTION

Prediction of complex phenotypic traits, particularly for complex diseases like Alzheimer's disease (AD) in humans, has seen increased traction in genomics research[1]. Over the past decade, genome-wide association studies (GWAS) have dominated genetic research for complex diseases like AD. Different approaches[2–4] like polygenic risk score (PRS) and wide-range of linear models have been proposed for risk prediction of complex diseases based on the genotype-phenotype associations for variants identified by GWAS. More recently, with increased data availability, non-linear methods like deep-learning[5] have been considered for constructing prediction models[6].

Most genetic studies of complex human traits has been undertaken in homogeneous populations from the same ancestry group[7], with the majority of studies focusing on European ancestry. For instance, despite African American individuals being twice as likely to develop AD compared to European, the genetic studies in African Americans are scarce[8]. Moreover, the genomic studies for admixed individuals are limited, although they make up more than one-third of the US population, with the population becoming increasingly mixed over time[9]. This lack of representation for minority populations and admixed individuals, if not mitigated, will limit our understanding of true genotype-phenotype associations and subsequently the development of genetic risk predictions. This will eventually hinder the long-awaited promise to develop precision medicine.

The discussion for mitigating the limited diversity in genetic studies has been more pronounced recently due to the consistent observation that the existing models have far greater predictive value in individuals of European descent than of other ancestries[10]. For example, PRS for blood pressure, constructed using GWAS of European ancestry, didn't generalize well for Hispanics/Latinos[11]. Similarly, the PAGE study found that genetic risk prediction models derived from European GWAS are unreliable when applied to other ethnic groups[12]. This lack of generalization to minority populations, including admixed individuals are critical limitations to human genetics. One major reason for this poor generalizability is the prediction models being confounded by underlying population structure. Both linear and non-linear models are susceptible to overfitting the training participants, which essentially comprises European ancestry in genomics. Although the non-linear models via deep learning produce impressive results across domains, they are more prone to overfitting – failing to generalize even with minor shifts in the training paradigm[13].

Studies have shown that one way to address this is by having multi-ethnic training participants. For instance, limited generalization of PRS for blood pressure, as reviewed above, substantially increased when GWAS of Europeans were combined with GWAS of Hispanics/Latinos[11]. Similarly, Martin et al. pointed out that the misdiagnoses of multiple individuals with African ancestry would have been corrected with the inclusion of even a small number of African Americans[14]. Although efforts to increase the non-European proportion of GWAS participants are being implemented, the proportion of individuals with African and Hispanic/Latino ancestry in GWAS has remained essentially unchanged[15]. As such, increasing training participants of other ancestries, particularly for admixed individuals, is not likely to occur any time soon without a dramatic priority shift, given the current imbalance and stalled diversifying progress over the recent years[10].

In this work, we aim to learn robust risk prediction models that generalize across different ancestry. Previous studies for learning robust and unbiased genotype-phenotype relationships against population bias have mostly been carried out with the PRS and its variants[16,17]. These studies are focused on specific ancestry or with an assumption that at least part of the background of the genome is still of European origin[17]. In addition, PRS won't capture the complex genotype-phenotype relationship. Alternatively, deep learning has recently shown improved predictability across domains (e.g., vision, language, etc.) due to its ability to capture the complex input-output relationship. However, standard deep learning approaches often fail to learn robust and unbiased representation, which is also the case for PRS-based methods. The major line of work for learning robust and unbiased representation with deep learning involves learning domain-invariant representation, where participants from different domains share common traits (e.g., genotype participants from different ancestry with common phenotype)[18,19]. Adversarial learning paradigms are often considered to learn such domain-invariant representations. However, such an adversarial approach is difficult to train and would require extensive hyperparameter tuning.

Here, we introduce DisPred, a novel deep-learning-based framework that can integrate data from diverse population to improve generalizability of genetic risk prediction. The proposed method combines a disentangling approach to separate the effect of ancestry from the phenotype-specific representation, and an ensemble modeling approach to combine the predictions from disentangled latent representation and original data. Unlike PRS-based methods, DisPred captures nonlinear genotype-phenotype relationships without restricting specific ancestral composition. Unlike adversarial learning-based methods, DisPred explicitly removes the ancestral effect from phenotype-specific representation and involves minimal hyperparameters tuning. Moreover, DisPred does not require self-reported ancestry information for predicting future individuals, making it suitable for practical use because human genetics literature has often questioned the definition and the use of an individual's ancestry[27,28], and at the minimum, the ancestry information may not be available during the test time. We evaluate DisPred performance to predict AD risk prediction in a multi-ethnic cohort composed of AD cases and controls and show that DisPred performs better than existing models in minority populations, particularly for admixed individuals.

## RESULTS

### Overview of the proposed workflow with DisPred

DisPred is a three-stage method to improve the phenotype prediction from the genotype dosage data (each feature has a value between 0 and 2). We present the workflow summary in **Figure 1**. First, as shown in **Figure 1 (A)**, we built a *disentangling autoencoder*, a deep-learning-based autoencoder, to learn phenotype-specific representations. The proposed deep learning architecture involves separating latent representation into ancestry-specific representation and phenotype-specific representation. In this way, this stage explicitly separates the ancestral effect from phenotype-specific representation. Second, as shown in **Figure 1 (B)**, we used the learned phenotype-specific representation extracted from the disentangling autoencoder to train the *prediction model*. We consider a linear model for the phenotype prediction model in our case. Since the phenotype-specific representation is nonlinear, the prediction model, despite being linear, will still capture the nonlinear genotype-phenotype relationship. Finally, as shown in **Figure 1 (C)**, we create *ensemble models* by combining the predictions from the learned representations, i.e., the result of our second stage, with the predictions from the original data, i.e., the existing approach of building prediction models. The second stage builds the prediction model from the disentangled representations and the ensemble in the third stage aims to enhance the prediction accuracy.

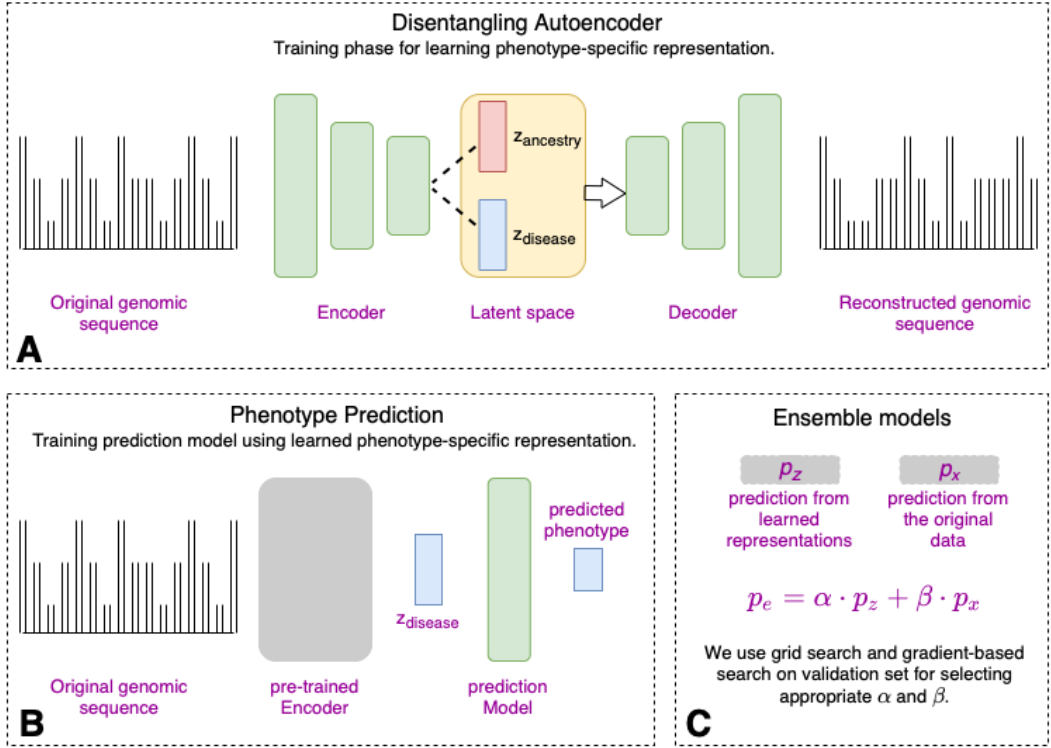

*Figure 1|Modeling strategy. (A) Disentangling autoencoder learns ancestry and phenotype-specific representation separately using autoencoder architecture and novel contrastive latent loss. (B) The disentangled phenotype-specific representation is then used for the phenotype prediction. A separate linear prediction model is trained on the obtained representation for the phenotype prediction. (C) To increase prediction power, we use the ensemble modeling approach, where the parameters can be obtained either from grid search or gradient-based search.*

We consider a training data $\mathcal{D} = \{\mathbf{x_i}, \mathbf{y_i}, \mathbf{a_i}\}_{i=1}^{N}$ with $N$ participants, where $\mathbf{y}$ represents the disease label (*e.g.,* case and control for binarized data), and $\mathbf{a}$ represents the environment label for the data (*e.g.,* categorical ancestral label). An encoder function $\mathcal{F}_\theta(\mathbf{x})$ decomposes original data $\mathbf{x}$ into ancestry-specific representation $\mathbf{z_a}$ and phenotype-specific representation $\mathbf{z_d}$, and a decoder function $\mathcal{G}_{\theta'}(\mathbf{z_a}, \mathbf{z_d})$ reconstruct the original data as $\hat{\mathbf{x}}$ using $\mathbf{z_a}$ and $\mathbf{z_d}$. Both the encoder function (θ) and decoder function (θ′) are parameterized by deep neural networks, and together represent the disentangling autoencoder. Primarily, the parameters $\theta$ and $\theta'$ are optimized by minimizing $\mathcal{L}^{Recon}$, the average reconstruction loss over $N$ training examples.

To disentangle the latent bottleneck representation, we propose a latent loss based on the following assumptions: for any data pair originating from the same environment, the corresponding pair of latent variables $\mathbf{z_a}$ should be similar, and different if data pair belong to different environment $\mathbf{a}$. Similarly, the latent $\mathbf{z_d}$ should be similar or different if the pair belong to the same or different disease label $\mathbf{y}$. We propose $\mathcal{L}^{SC}$, contrastive loss[20–22] to enforce these similarities in the latent space. We apply $\mathcal{L}^{SC}$ independently to both $\mathbf{z_a}$ and $\mathbf{z_d}$. Overall, the objective function for training the disentangled autoencoder takes the following form:

$$\mathcal{L}^{Disentgl-AE} = \mathcal{L}^{Recon} + \alpha_d \cdot \mathcal{L}^{SC}_{z_d} + \alpha_a \cdot \mathcal{L}^{SC}_{z_a}$$

where $\alpha_*$ represent the hyperparameter for the corresponding latent loss. For each latent variable, the contrastive loss will enforce the encoder to give closely aligned representations to all the entries from the same label in the given batch encouraging disentanglement of disease and environment features in two separate latent variables.

In the second stage, we utilize the trained autoencoder to extract phenotype-specific representation $\mathbf{z_d}$ to train prediction model for the given phenotype. Our prediction models are trained with linear regression, regressing $\mathbf{z_d}$ to the corresponding disease label $\mathbf{y}$. Finally, in the third stage, we create ensemble models by combining $p_z$, the predictions from the linear model using the phenotype-specific representation $\mathbf{z_d}$, with $p_x$, the predictions from the linear model using the original data $\mathbf{x}$:

$$p_e = \alpha \cdot p_z + \beta \cdot p_x$$

where $\alpha$ and $\beta$ are the weighing parameters determined using gradient-based search using the validation set. We use the same training and validation data as in earlier stages to obtain $p_z$ and $p_x$ and learn the weighing parameters $\alpha$ and $\beta$. Here $p_e$ represents prediction from additive ensemble model, a complex model that combines the predictions resulting from linear and nonlinear relationship between original data and the phenotype target.

**Application of DisPred to Alzheimer's Disease (AD) risk prediction**

We use the DisPred framework to predict AD using genetic data. The aim is to evaluate the prediction accuracy (via Area under the curve: AUC) of the proposed DisPred and Disentangling Autoencoder (Disentgl-AE) in comparison to other conventional methods in genomics trained on participants from European ancestry, including Polygenic Risk Score (PRS), linear models (Lasso), and the supervised Neural Network (NN). Although unconventional, we also consider the models trained using other non-European ancestries like African American population (AFR) to understand the effect of training models with training participants other than European ancestry. Furthermore, we also considered adversarial learning to capture domain invariant representations where we treated ancestry as domains (Adv). We applied these methods to two cohorts: the Alzheimer's Disease Sequencing Project (ADSP) and the UK Biobank (UKB). We considered genetic variants identified by existing GWAS as features, including 5,014 variants associated with AD from Jansen et al., 2019[23] (variants with *p < 1e-5*) and 78 variants from Andrews et al., 2020[24]. The outcome or phenotype label for the ADSP is the clinical diagnosis for the presence or absence of AD, and for the UKB is dichotomized version of continuous proxy-phenotype derived from the information from family history (first-degree relative with reported AD or dementia). After filtering for participants and variants, the final dataset includes 11,640 participants with 3,892 variants for the ADSP cohort and 461,579 participants with 4,967 variants for the UKB.

We split the data into training and test and further divided the training set into training and validation sets with stratification based on the phenotype labels. We consider self-reported ancestry labels in the training dataset to train the disentangling autoencoder and other ancestry-specific linear and non-linear models. The independent test data consists of 2,101 participants for the ADSP and 138,474 participants for the UKB. To evaluate the performance of the proposed method to predict AD status in minority populations or admixed individuals, we estimated ancestry percentages from genome-wide data using SNPWeight v2.1[25]. Then we used these ancestry estimates to divide the test data into different ancestries to present the results. This is to ensure that the partitions accurately reflect the actual genetic-ancestry background and enable a more rigorous evaluation of the methods. Details for dataset preparation are explained in the Methods section.

**Representation learned via disentangling autoencoder**

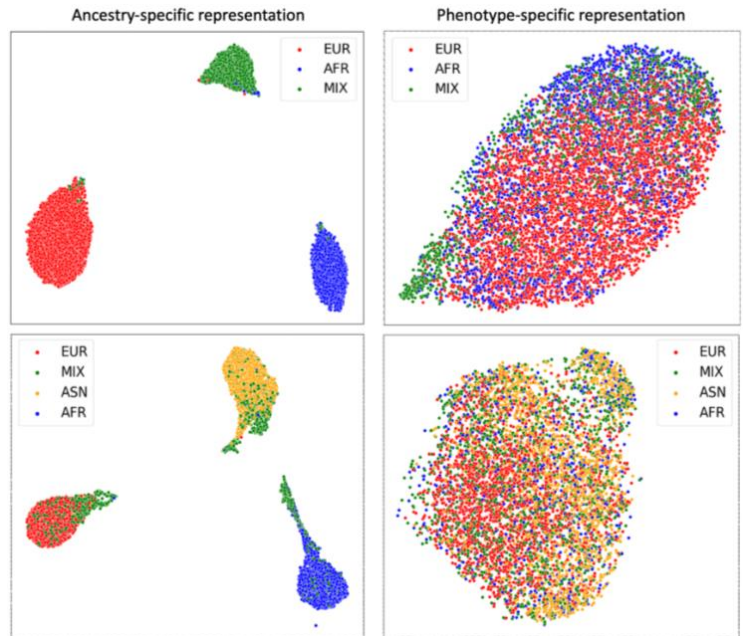

*Figure 2 UMAP plots of ancestry-specific representation (left column) and phenotype-specific representation (right column) learned from Disentgl-AE for the ADSP (top row) and the UKB (bottom row). We colored the points with self-reported ancestry labels (EUR: European, MIX: admixed, ASN: Asian, and AFR: African), demonstrating that phenotype-specific representation is invariant to the ancestry background.*

We analyzed the representations learned from the proposed DisPred framework. In **Figure 2**, we present the Uniform Manifold Approximation and Projection (UMAP)[26] plots for the ancestry-specific representation $z_a$ and phenotype-specific representation $z_d$ to assess whether the proposed method can separate the ancestry effect from the GWAS variants. First, we note that the representation captured the ancestry-related information on the left plots. The three training ancestry groups for the ADSP cohort were clearly separated and three out of four training ancestry groups for the UKB cohort were also separated. However, the admixed (MIX) group in the UKB cohort is mixed with other ancestries (left panel on bottom row). This is a limitation of the proposed regularization, or the proposed architecture, which may not be able to completely separate the MIX group from other ancestries. On the right side of both figures, when ancestry labels are applied to the phenotype-specific representation $z_d$, we note that the ancestry labels are scattered without forming clear clusters. Meaning that our proposed method correctly identified a latent representation $z_d$ that is invariant to ancestry background, which we later use to build the prediction models.

# DisPred improved AD risk prediction for the minority populations and for admixed individuals

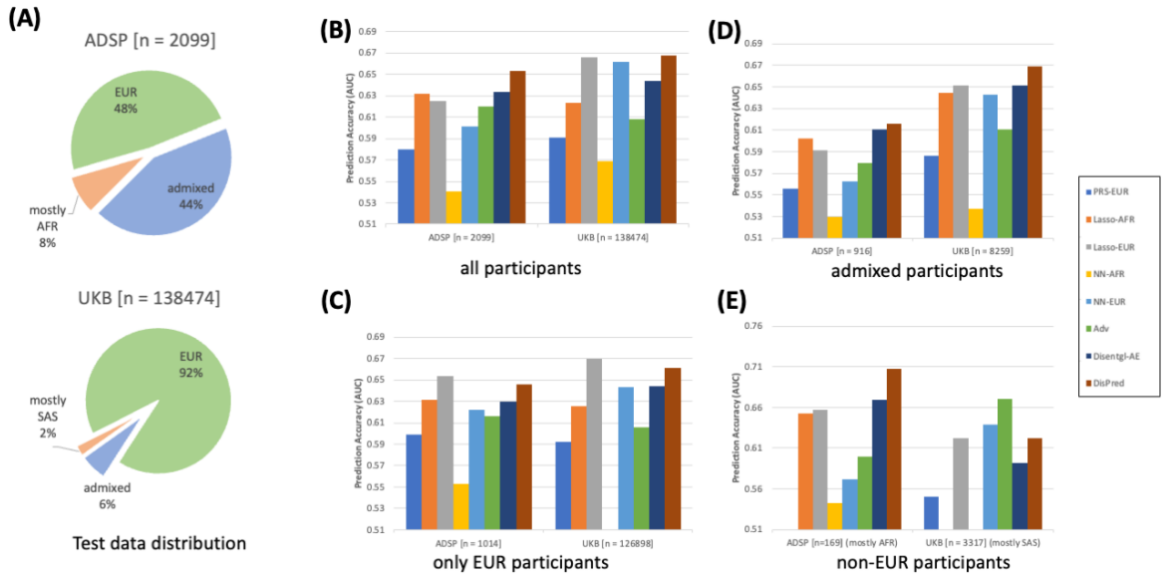

*Figure 3* Test data distribution for the ADSP and the UKB cohorts (A). In the pie chart, we present the proportion of admixed participants compared to the European participants and other dominant non-European participants, i.e., African American (AFR) for the ADSP and South Asian (SAS) for the UKB. Prediction accuracy (AUC) for different models (PRS trained on European (PRS-EUR), Lasso trained on African American (Lasso-AFR), Lasso trained on European (Lasso-EUR), Neural Network trained on African American (NN-AFR), Neural Network trained on European (NN-EUR), Adversarial Neural Network trained on all data (Adv), Disentangling Autoencoder (Disentgl-AE), and DisPred trained on all data) on different subsets of the test dataset: all participants (B), EUR participants (C), admixed participants (D), and participants from non-European ancestries (E) in the test dataset for the ADSP and the UKB cohorts.

We report the main results in **Figure 3**. First, test data distribution for different ancestries for the two cohorts is presented in panel A. We considered 90% and 65% as estimated ancestry cut-offs, respectively, for the ADSP and the UKB to stratify test participants into five super populations: South Asians (SAS), East Asians (EAS), Americans (AMR), Africans (AFR), and Europeans (EUR), and an admixed group composed of individuals not passing cut-off in any ancestry. Since UKB primarily comprises European participants, we set the threshold low to obtain enough admixed and non-EUR participants for our analysis. In both datasets, EUR ancestry individuals correspond to the largest proportion (48% for the ADSP and 92% for the UKB). Admixed individuals are the second largest group (44% for the ADSP and 6% for the UKB). For both cohorts, we consider four test datasets: i) all test participants, ii) European participants, iii) admixed participants, and iv) non-European participants, primarily AFR for the ADSP and primarily SAS for the UKB. Panel B presents the result for all the test participants, and the proposed DisPred leads to the best result for ADSP and the joint best result for UKB (Lasso-EUR being marginally better than DisPred). The analysis restricted to European participants is shown in panel C. In this case, Lasso-EUR achieves the best result, and the proposed method produces the second-best result for both datasets. In admixed individuals, DisPred produces the best result for both UKB and ADSP (Panel D). In UKB, the performance on admixed individuals (best AUC: 0.6692 by DisPred) is equivalent to the one obtained in the EUR individuals (best AUC: 0.6695 by Lasso-EUR), but for the ADSP, the predictive ability is better in EUR (best AUC: 0.6532 by Lasso-EUR) compared to admixed individuals (best AUC: 0.6157 by DisPred). Increased heterogeneity in the composition of admixed individuals in the ADSP cohort may explain this difference. The top two ancestral composition for the ADSP's admixed group includes 45.31% EUR and 40.72% AFR, compared to 50.51% EUR and 28.50% AFR for the UKB. The result for non-European population is shown in panel E. For AFR participants in ADSP, DisPred again produces the best result. However, for the SAS in UKB, the adversarial trained NN (Adv) leads to the best AUC. Except for this group, in general, Adv demonstrates poor generalizability. Similarly, the predictive ability of PRS is substantially lower in non-European participants. On the other hand, Lasso-based models, especially trained on EUR data, performed competitively, especially in the groups dominated by European ancestry. Compared to these linear and non-linear models, the proposed Disentgl-AE utilizes all the available training participants to separate the effect of ancestry from the phenotype representation. As such, the obtained phenotype representation demonstrated the best predictive abilities.

**DisPred performs better in the presence of ancestral mismatch and dataset shift**

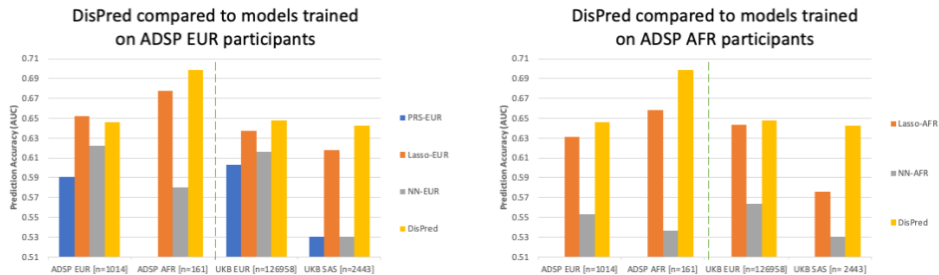

*Figure 4* Prediction accuracy (AUC) for different models (PRS trained on European (PRS-EUR), Lasso trained on European (Lasso-EUR), Lasso trained on African American (Lasso-AFR), Neural Network trained on European (NN-EUR), Neural Network trained on African American (NN-AFR), and DisPred) on the practical setting of when there is alignment or mismatch between the ancestry (left side of dotted lines) or dataset shift (right side of dotted lines) of training and test participants. All prediction models are trained on ADSP (EUR participants on the left and AFR participants on the right).

In a practical setting, since existing methods are often ancestry specific, ancestry information is required for predicting future patients or individuals. However, human genetics literature has often questioned the definition and the use of an individual's ancestry[27,28], and at the minimum, the ancestry information may not be available during the test time. Without accurate ancestry information, the participants could either align or mismatch with the corresponding ancestry of training populations. Similarly, dataset shift (or distribution shift) might exist in training and test participant, i.e., application of models trained on one cohort to participants of different cohorts. For such conditions, we present the results in **Figure 4**. We consider four different test scenarios: i) ADSP EUR (n=1,014), ii) ADSP AFR (n=161), iii) UKB EUR (n=126,958), and iv) UKB SAS (n=2,443), and compared the performance of DisPred against models trained with ADSP EUR participants (left panel), and models trained with ADSP AFR participants (right panel). We use the same ancestry percentage cut-off, as used in the previous section, to stratify the test participants into EUR and AFR for ADSP and EUR and SAS for UKB. Since the effect sizes (or betas) for calculating PRS are derived from EUR participants, we don't have results for PRS-AFR in the above table. In each panel, the dotted lines separate the scenarios for ancestral alignment or mismatch (left) and the dataset shift (right). We note that among all the analyzed cases, only when the EUR test participants aligned with the EUR training population, DisPred obtain less predictive accuracy than Lasso. Other than that, DisPred achieved better results when the test participants didn't align with the training population for AFR and EUR analysis and for AFR, even when they aligned. Furthermore, in the case of dataset shift, DisPred performs better in all the cases, including when models trained on EUR participants (ADSP) are applied to the EUR participants (UKB). We also note that although we did not leverage data from SAS, we significantly improved prediction accuracy for UKB SAS. This demonstrates that we learned an invariant representation robust to ancestry background. Overall, these results show the practical usability of the DisPred model over existing methods when there is an ancestral mismatch or dataset shift.

**DisPred with increasing individual-level heterogeneity**

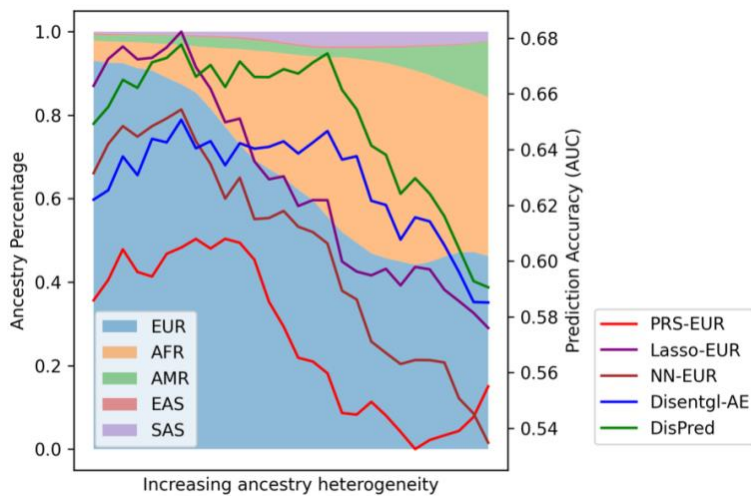

*Figure 5 Prediction accuracy (AUC) for different models on data subsets with increasing ancestry heterogeneity for the ADSP cohort. The background is color-coded with the proportion of self-reported ancestry (EUR: European, AFR: African American, AMR: American, EAS: East Asian, and SAS: South Asian) for each data subset.*

DisPred, compared to existing methods, improved predictability for the minority population, particularly for admixed individuals. Since admixed individuals' genomes is an admixture of genomes from more than one ancestral population, we identified that different models struggle as individual heterogeneity increases. This section presents an in-depth evaluation of this issue for the ADSP cohort. Estimated ancestral percentages was used to calculate the heterogeneity. For each sample in the test set, the variance of ancestral proportion was computed, and we sorted these in decreasing order, obtaining a sequence from homogeneity to heterogeneity. For such arrangement, **Figure 5** shows the results, where we created numerous data subsets by sliding window-based process with a window size of 750 participants and stride length of 50 participants. This way, the data subset heterogeneity increases from left to right. The background is color-coded with the proportion of self-reported ancestry for each data subset. All methods considered in this work start to degrade as we move toward heterogeneity or admixed individuals. The proposed DisPred performs well when there is a sharp decline in the proportion of European ancestry (in the middle region of the graph) and slowly decreases towards the end. Overall, the DisPred produces the best result. These results also suggest that the lack of generalization cannot be addressed by simply increasing the non-European training participants by recruiting more homogeneous individuals from minority populations. An equal effort needs to be given toward designing unbiased and fairer algorithms like DisPred.

## DISCUSSION

As the population becomes increasingly mixed over time, understanding how to analyze and interpret admixed genomes will be critical for enabling transethnic and multiethnic medical genetic studies and ensuring that genetics research findings are broadly applicable. Yet, existing AI-based approaches in genomics are largely focused on homogeneous European ancestry. We thus urge a joint research effort to confront the existing approach for ancestry-specific AI frameworks and focus on building unbiased alternatives. This study introduced DisPred, a deep learning-based framework for improving AD risk prediction in minority populations, particularly for admixed individuals. First, we developed a disentangling autoencoder to disentangle genotype into ancestry-specific and phenotype-specific representations. We then build predictive models using only phenotype-specific representation. Finally, we used ensemble modeling to combine the prediction models built using disentangled latent representation, and the model built using the original data. We constructed DisPred for unbiased and robust predictions for diverse ancestry, particularly for the non-European population. With AD GWAS data from the ADSP and the UKB cohort, we confirmed the effectiveness of the disentanglement-based framework for admixed individuals.

In this study, we also demonstrated that, unlike existing practices where models built for particular ancestry are applied to the individuals of the same ancestry requiring ancestry information at the test time, DisPred predicts the risk of test individuals without any assumption on their ancestry information. We believe this is an appealing feature of DisPred as there has been a long-standing debate on the use of self-reported race/ethnicity/genetic ancestry in biomedical research. The scientific, social, and cultural considerations make it challenging to provide an optimal label for race or ethnicity, which could result in ambiguity, contributing to misdiagnosis. Moreover, at the minimum, such ancestry information may not be available. As such, a method like DisPred that does not require ancestry information is beneficial to get around this critical issue. Similarly, dataset shift, i.e., different training and test data distributions (or cohorts), although a common practical scenario, makes generalization challenging for machine learning models. We showed that compared to existing methods, DisPred demonstrated better predictive abilities in the presence of dataset shift. Overall, existing methods typically improve predictions by leveraging data from the target population and, in turn, make the model more sensitive to the ancestry background and thus require relevant ancestry information at prediction stages. Unlike them, DisPred is different and novel as it learns invariant representation robust to ancestry background and performs well in a diverse test environment.

However, certain limitations are not well addressed in this study. First, in this study, we didn't perform any feature importance analysis to understand if specific variants were more pronounced for admixed groups than the homogenous European ancestry. Due to the multi-stage framework involving different training and optimization processes, unlike other AI approaches, it is not straightforward to relate the predicted phenotype to the original data level. A detailed study is required to trace the flow of information from the genetic variants to disentangled representations and then to the predicted phenotype. The second is the engineering efforts. The deep learning architecture, proposed in this work constructed using multi-layer perceptron with ReLU activation, potentially has space for improvements. Our framework used contrastive loss[20–22] to achieve disentanglement in the latent space. In recent years, many advanced similarity-enforcing losses[29,30] have been proposed in the deep learning literature, which could improve our framework. We have open-sourced the disentangling autoencoder to encourage future research in this direction. It is worth mentioning that our proposed framework is not confined to the AD risk prediction and can be extended to other phenotypes.

**MATERIALS AND METHODS**

In this section, we first describe how we constructed the dataset and provide details of our proposed method and implementations for collaboratively training the AI model.

**Dataset preparation**

In this study, we consider two datasets: the ADSP and the UKB. We use the $p<1e-5$ threshold to obtain the candidate regions of 5,014 GWAS variants, obtained from Jansen et al., 2019[23] and Andrews et al., 2020[24]. We remove SNPs with more than a 10% missing rate to ensure marker quality. We then remove participants with absent AD phenotype or ancestral information. For the ADSP, we have a dichotomous case and control label for AD phenotype. For the UKB, we use the AD-proxy score defined in Jansen et al., 2019[23], which combines the self-reported parental AD status and the individual AD status. Since AD is an age-related disease, we removed control participants below age 65 (age-at-last-visit). For the ADSP, since case and controls are well defined, we perform this step for the whole dataset, but for the UKB, we first split data into training and test and remove AD proxy score less than 2.0 below age 65. A total of 10,504 participants with 3,892 variants are obtained for the ADSP cohort and 461,579 participants with 4,967 variants for the UKB. We held out 20% and 30% of the participants, respectively, for the ADSP and UKB, for the test data, and the rest as the training data. The training data consists of 8,403 participants, including 2,061 AFR, 4,780 EUR, and 1,562 HIS, for ADSP and 323,105 participants, including 4,493 AFR, 306,169 EUR, 5,406 HIS, and 7,037 ASN for UKB based on self-reported ancestry labels. These self-reported ancestry labels are incorporated into the proposed method to learn the latent representation. Further, we held out 1,000 participants for the ADSP and 10,000 participants for the UKB as the validation data. The data division into training, test, and validation is stratified using phenotype labels, i.e., each set contains approximately the same percentage of participants of phenotype labels as the complete set. For the UKB, we dichotomized the AD-proxy score with a threshold of 2.0. Overall, the number of training, test, and validation participants for the ADSP are 7,403, 2,101, and 1,000 and for the UKB are 313,105, 138,474, and 10,000, respectively.

**Ancestry determination**

For each cohort included in our analysis, we first determined the ancestry of each individual with SNPWeight v2.1[25] using reference populations from the 1000 Genomes Consortium[31]. Prior to ancestry determination, variants were filtered based on genotyping rate (< 95%), minor allele frequency (MAF < 1%) and Hardy-Weinberg equilibrium (HWE) in controls ($p < 1e-5$). By applying an ancestry percentage cut-off > 90% (for ADSP) and > 65% (for UKB), the participants were stratified into five super populations: South Asians, East Asians, Americans, Africans, and Europeans, and an admixed group composed of individuals not passing cut-off in any ancestry.

**Disentangling Autoencoder**

The disentangling autoencoder comprises of an encoder function $\mathcal{F}_\theta(\cdot)$ and the decoder function $\mathcal{G}_{\theta'}(\cdot)$. The encoder decomposes original data $\mathbf{x}$ into ancestry-specific representation $\mathbf{z_a}$ and phenotype-specific representation $\mathbf{z_d}$, and the decoder reconstruct the original data as $\hat{\mathbf{x}}$ using $\mathbf{z_a}$ and $\mathbf{z_d}$, which are concatenated together. The parameters of the autoencoder, *i.e.,* $\theta$ and $\theta'$, are optimized by minimizing the $\mathcal{L}^{Recon}$, average reconstruction error over $N$ training examples.

$$\mathcal{L}^{Recon} = \frac{1}{N}\sum_{i=1}^{N} L_r(x_i, \hat{x}_i) = \frac{1}{N}\sum_{i=1}^{N} L_r\left(x_i, \mathcal{G}_{\theta'}(\mathcal{F}_\theta(x_i))\right)$$

where $L_r$ is mean squared error. To achieve disentanglement, we propose contrastive loss[20–22] to enforce the similarities between the latent representations obtained from the data pair in the latent space. For a batch with $N$ randomly sampled pairs, $\{x_k, y_k\}_{k=1}^{N}$, the contrastive learning algorithm use two random augmentations (commonly referred to as ``view'') to create $2N$ pairs $\{\tilde{x}_l, \tilde{y}_l\}_{l=1}^{2N}$, such that $\tilde{y}_{[1:N]} = \tilde{y}_{[N+1:2N]} = y_{[1:N]}$. In our case, we simply replicate the batch to create $2N$ pairs. Using this multi-viewed batch, with index $i \in I \equiv \{1 \dots 2N\}$ and its augmented pair $j(i)$, the supervised contrastive (SC) loss[32] takes the following form:

$$\mathcal{L}^{SC} = -\sum_{i \in I} \frac{1}{S(i)} \sum_{s \in S(i)} \log \frac{\exp(\mathbf{z}_i \cdot \mathbf{z}_s)/\tau}{\sum_{r \in I \setminus i} \exp(\mathbf{z}_i \cdot \mathbf{z}_r)/\tau}$$

where $S(i) \equiv \{s \in I \setminus i : \tilde{y}_p = y_p\}$ is the set of indices of all the positives in the multi-viewed batch distinct from $i$ and $\tau$ is a temperature parameter. We apply $\mathcal{L}^{SC}$ independently to both $\mathbf{z_a}$ and $\mathbf{z_d}$. Overall, the objective function for training disentangled autoencoder takes the following form:

$$\mathcal{L}^{Disentgl-AE} = \mathcal{L}^{Recon} + \alpha_d \cdot \mathcal{L}^{SC}_{z_d} + \alpha_a \cdot \mathcal{L}^{SC}_{z_a}$$

where $\alpha_*$ represent the hyperparameter for the corresponding latent loss. We set these hyperparameters through grid-search using held-out validation set. For each latent variable, the contrastive loss will enforce encoder to give closely aligned representations to all the entries from the same label in the given batch encouraging disentanglement of disease and environment features in two separate latent variables.

**Model architecture and training setting**

We used the Python package scikit-learn[33] to implement all the linear models and PyTorch[34] to implement all the non-linear models, including the proposed Disentangling Autoencoder. The proposed disentangling autoencoder (**Figure S1**, top row) comprises four layers in the encoder and three layers in the decoder. The encoder layers are designed to learn low-dimensional features, and the decoder layers to upscale the low-dimensional features into high-dimensional data space. We consider the ReLU activation function for all the non-linear layers in the autoencoder. We train DisentglAE with Adam[35] optimizer with a constant learning rate of 5e-3. We train the model till $N$ epochs. We first let the model focus entirely on the reconstruction task till $N_1$ epochs by setting the weight parameters for similarity loss as 0. From $N_1$ + 1 epochs and to $N_2$ epochs, we linearly ramp the weight parameters $\alpha_*$ to 1, and then continue training till $N$ epochs. We conduct hyperparameter testing selecting the model that results in the best validation AUC for the following hyperparameters: $\mathbf{z_d}$ (30, 40, 50), $\mathbf{z_a}$ (30, 40, 50), $\tau$ (0.03, 0.05), and $\alpha_*$ (0.0001, 0.0003). For the ADSP, $\mathbf{z_d}$, $\mathbf{z_a}$, $\tau$, and $\alpha_*$ are selected respectively as 40, 40, 0.03 and 0.0001, and for the UKB, $\mathbf{z_d}$, $\mathbf{z_a}$, $\tau$, and $\alpha_*$ are selected respectively as 40, 30, 0.05 and 0.0001. Other used values for the hyperparameter include, for the ADSP, $N$ = 500, $N_1$ = 100, $N_2$ = 250, and batch size = 256. For the UKB, $N$ = 100, $N_1$ = 10, $N_2$ = 70, and batch size = 256. We then use ordinary least square linear regression to minimize the residual sum of squares between $\mathbf{y}$, the phenotype labels and $\boldsymbol{p_z}$, the targets produced by the $\mathbf{z_d}$.

For the ensemble modeling, we combine $\boldsymbol{p_z}$, predictions from learned representations and $\boldsymbol{p_x}$, prediction from the original data. For the prediction from the original data, we consider Lasso linear model to make fair comparison with existing models. To combine these predictions, we conduct both grid-search and gradient-based search. For grid-search, we test all the values from 0.1 to 1.5, increasing at 0.1 for both $\alpha$ and $\beta$, and found $\alpha$ = 1.4 and $\beta$ = 0.4 for the ADSP, and $\alpha$ = 1.2 and $\beta$ = 0.6 for the UKB, that produces the best AUC for the validation set. For the gradient-based search, we consider $\alpha$ and $\beta$ as the parameter, initialize their weights as 1.1 and 0.9, and train the ensemble function ($p_e$) with an SGD optimizer for 5000 epochs. The best result was produced by $\alpha$ = 1.103 and $\beta$ = 0.720 for the ADSP, and $\alpha$ = 1.121 and $\beta$ = 0.560 for the UKB. The reported results for both UKB and the ADSP are using the parameters from grid-search. The result from the gradient search was similar to the grid-search result.

**Baseline methods**

In **Figure S1** (middle row and bottom row), we present the architecture for other neural network-based methods considered in this work. This includes supervised Neural Network (NN) and adversarial learning for capturing domain invariant representations with ancestry as domains (Adv). For training NN, we consider following parameters for ADSP: number of epochs = 200, learning rate = 5e-3, batch size = 64, and for the UKB: number of epoch = 100, learning rate = 5e-3, batch size = 256. For Adv, we consider Wasserstein distance for capturing domain invariant representation, and followed the experimental setup from Shen et al., 2017[36].

The Lasso linear models are fitted with iterative fitting along a regularization path and the best model is selected by 5-fold cross-validation using whole training set. We set alphas automatically and consider $N_{alpha}$ (number of alphas along the regularization path) = 10, maximum number of iterations = 5000 and tolerance value for optimization = 1e-3. For the polygeneic risk score (PRS), we follow the standard approach by computing the sum of risk alleles corresponding to the AD phenotype for each sample, weighted by the effect size estimate of the most powerful GWAS on the phenotype. We obtain the effect size from the genetic variants identified by Jansen et al., 2019[23] and Andrews et al., 2020[24].

## Acknowledgements
**Funding:** This work is supported by NIH/NIA award AG066206 (ZH).

## Author Contributions
P.K.G, J.Z, and Z.H. developed the concepts for the manuscript and proposed the method. P.K.G, Y.L.G, X.L., J.Z, and Z.H. designed the analyses and applications and discussed the results. P.K.G, J.Z, and Z.H. conducted the analyses. Z.H., J.Z, and H.T. helped interpret the obtained results.

P.K.G, Z.H., J.Z., Y.L.G, X.L., and H.T. prepared the manuscript and contributed to editing the paper.

**Competing interests:** The authors declare no competing interests.

**Data and materials availability:** The dataset used in this paper i.e., the Alzheimer's Disease Sequencing Project (ADSP) and the UK Biobank (UKB) are publicly available data cohorts. We used publicly available summary statistics from Jansen et al., 2019[23] and Andrews et al., 2020[24]. The code for data preprocessing is written in R programming language. The code for DisPred's training, prediction and evaluation are written in Python with PyTorch and scikit-learn. The codes will be made freely available on the GitHub platform.

## Supplementary Materials

**Figure S1. Neural network architectures.** The top row represents architecture for Disentangling autoencoder (**DisentglAE**), the middle row represents architecture for supervised Neural Network (**NN**), and the bottom row represents Adversarial learning (**Adv**). For all the networks except Adv Critic (bottom row, right), the genotype data is presented to the network as the first input data with dimension (input dim) of 3892 for the ADSP and 4967 for the UKB. For Adv Critic, the input dimension is the output of the $FC_2$ layer of Adv Backbone. Linear Layer represents multi-layer perceptron, and ReLU represents Rectified Linear Unit, a nonlinear activation function. For DisentglAE, $z_d$ dim and $z_a$ dim represents latent dimensions respectively for phenotype-specific representation and ancestry-specific representation.

| Layer | DisentglAE |
|---|---|
| $FC_1$ | Linear Layer (input dim, 400), ReLU |
| $FC_2$ | Linear Layer (400, 200), ReLU |
| $FC_{31}$, $FC_{32}$ | Linear Layer(200, $z_d$ dim), Linear Layer(200, $z_a$ dim) |
| $FC_4$ | Linear Layer ($z_d + z_a$ dim, 200), ReLU |
| $FC_5$ | Linear Layer (200, 400), ReLU |
| $FC_6$ | Linear Layer (200, input dim), ReLU |

| Layer | NN |
|---|---|
| $FC_1$ | Linear Layer (input dim, 200), ReLU |
| $FC_2$ | Linear Layer (200, 100), ReLU |
| $FC_3$ | Linear Layer(100, 20), ReLU, dropout (p = 0.5) |
| $FC_4$ | Linear Layer(20, 20), ReLU, dropout (p = 0.5) |
| $FC_6$ | Linear Layer (20, 1) |

| Layer | Adv Backbone |
|---|---|
| $FC_1$ | Linear Layer (input dim, 200), ReLU |
| $FC_2$ | Linear Layer (200, 20), ReLU |
| $FC_3$ | Linear Layer (20, 1) |

| Layer | Adv Critic |
|---|---|
| $FC_1$ | Linear Layer (20, 10), ReLU |
| $FC_2$ | Linear Layer (20, 1) |